\documentclass[11pt,letterpaper]{article}
\usepackage{advdate}
\date{\AdvanceDate[-1600]\today}
\usepackage{hyperref}
\usepackage{mathptmx}
\usepackage[pdftex]{graphicx}
\usepackage{tabularx,ragged2e,booktabs,caption}

\usepackage[round]{natbib} 
\usepackage[usenames,dvipsnames]{xcolor}
\newcommand{\blue}[1]{\textcolor{RoyalBlue}{#1}}
\newcommand{\fillme}[1]{\blue{\texttt{[Insert #1]}}}
\newcommand{\instructions}[1]{\blue{\textit{#1}}}

\renewcommand{\instructions}[1]{}
\renewcommand{\fillme}[1]{}

\begin{document}
\title{Fairness in AI Systems: Mitigating gender bias from language-vision models\fillme{}}
\author{Lavisha Aggarwal,  Shruti Bhargava\fillme{}}
\maketitle

\instructions{CS546 involves a research project. This is a template
  for the initial proposal, which will be help us make sure your
  project is both suitable, feasible and interesting. 
You can just uncomment the following two lines in the preamble of the .tex file for the instructions to disappear:
\texttt{\%$\backslash$renewcommand\{$\backslash$instructions\}[1]\{\}}~ and \texttt{\%$\backslash$renewcommand\{$\backslash$fillme\}[1]\{\}}
}

\section*{1. Introduction}
Our society is plagued by several biases, including racial biases, caste biases, and gender bias. As a matter of fact, several years ago, most of these notions were unheard of. These are human creations that passed through generations, with much amplification leading to the scenario wherein several biases have taken the role of expected norms or required laws by certain groups in the society. One notable example is of gender bias. The notion about the role of a man, and that of a woman have taken different forms across the globe. Whether we talk about the corporate world, political scenarios or daily lifestyle, some generic differences are observed regarding the involvement of both the groups. This differential distribution, being a part of the society at large, exhibits its presence even in the recorded data. Now, the field of machine learning is completely dependent on the available data. The idea of learning from data and making predictions based on that assumes that the data defines the expected behaviour at large. This is a huge obstacle in the path for creating a world with equality and justice. This work studies and attempts to alleviate the gender bias issues from the task of image captioning.   
\section*{2. Motivation and Related Work}
The motivation for this project stems from the recent paper \cite{bias} highlighting the rampant gender bias severely infecting most of the present day machine learning and natural language based models. They experimented with word analogies and found that due to the inherent bias present in the training data, relations were developed between certain kind of activities or objects and gender words \textit{('man'- 'doctor', 'woman'-'homemaker' etc.)} leading to incorrect inferences. It is observed that this gender bias in datasets is a demonstration of societal behaviours, owing to the skewed data distributions in existing datasets, and also from the human annotators' prejudices, responsible for labelling the datasets [\cite{selvaraju2016grad}][\cite{learning}][\cite{flickr}][\cite{vqatest}]. These biased data result in biased models (for example the Word2Vec word embeddings) which are further used for more advanced models and tasks\cite{survey} \cite{yinyang}. 
\subsection*{2.1 Gender Bias via human annotation}
We manually analysed the MS-COCO dataset and found some disturbing ground-truth annotations for the captions. Several images, wherein there is a partially visible human who can't be distinguished as a man or woman, most human annotators use their bias about the expected (or more common) gender from the activity in the image. For instance, a human on a bike is labelled as 'man', since driving a bike is considered uncommon for women. Instead of  accepting the human as a gender neutral person, annotators tend to use their own bias. Following are selected examples where all human annotators labelled the human as male even though the gender is not clear from the image. There are many such instances, though we do not report it here due to space crunch. 
\begin{center}\includegraphics[height=4.8cm]{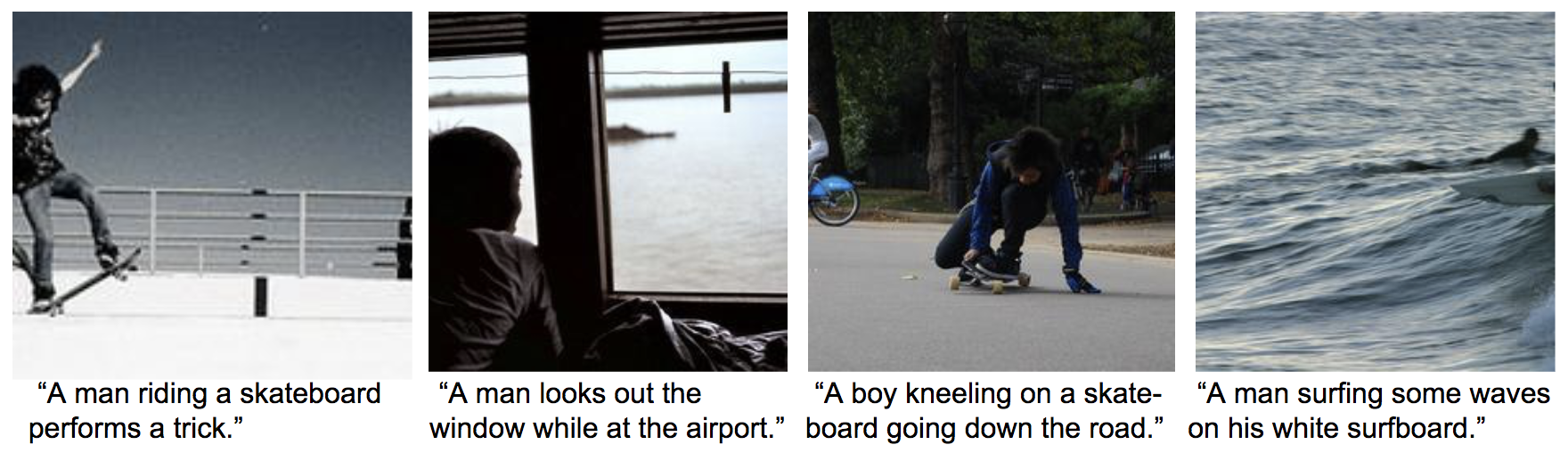}\end{center}
The dataset ground truth captions for these images are mentioned below each of them. The surprising point to note is that, for each of the images,  all 4 human annotators labelled the person as ‘man’ when it can be clearly seen that the gender is not distinguishable from the image. 
\subsection*{2.2 Gender Bias in the existing datasets}
In the figure below, we observe the gender bias present in some of the popular datasets, namely MS-COCO and imSitu Verb. It reflects that words like 'braiding', 'shopping', 'washing' occur much more frequently with females than males in the data whereas words like 'repairing' and 'coaching' have been seen much more with males.\cite{men} \cite{women} 
\begin{center}\includegraphics[height=6cm]{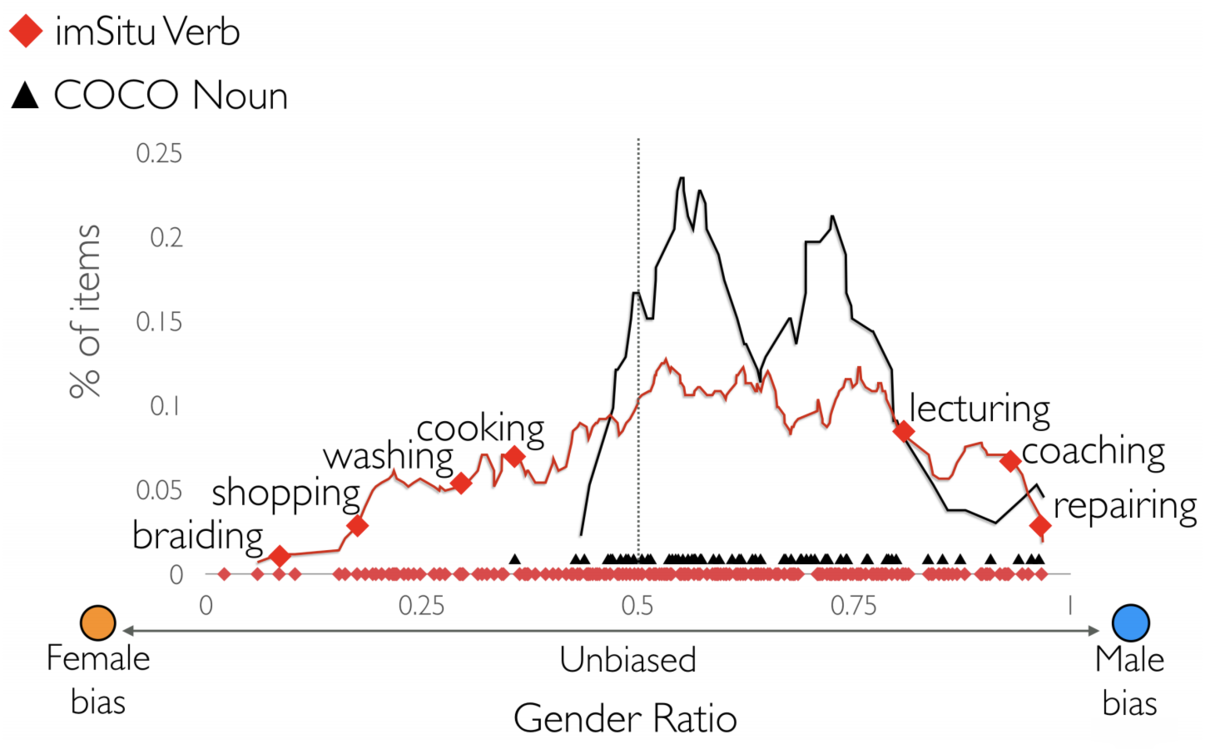}\end{center}

\subsection*{2.3 Gender Bias in some of the state-of-the-art models}
Next we analyse how this gender bias leads to the faulty performance of our systems relying on ML models. The figure below demonstrates Google Translate translating a text from the gender-neutral language Turkish to English, a gender aware language. It shows that the model assigns a male gender to a 'doctor' and a female gender to 'nurse' to an initially gender neutral sentence in Turkish.\begin{center}\includegraphics[height=3cm]{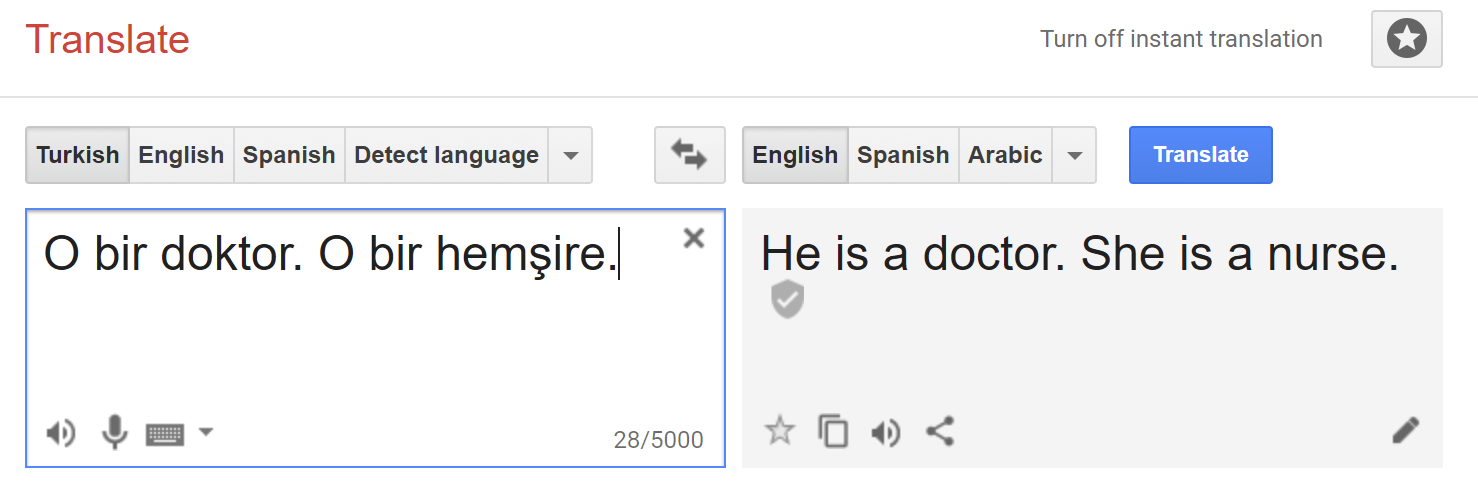}\end{center}
In another trained model for  image captioning, we observe below that the model labels the subject as 'man' giving the maximum attention to the computer screen. This again signifies the relation between a computer and man learnt inherently by the model from the training data. The image clearly contains a woman, but the  model instead uses the prior associated with computers to make the gender decision.[\cite{women} ] \begin{center}\includegraphics[height=4cm]{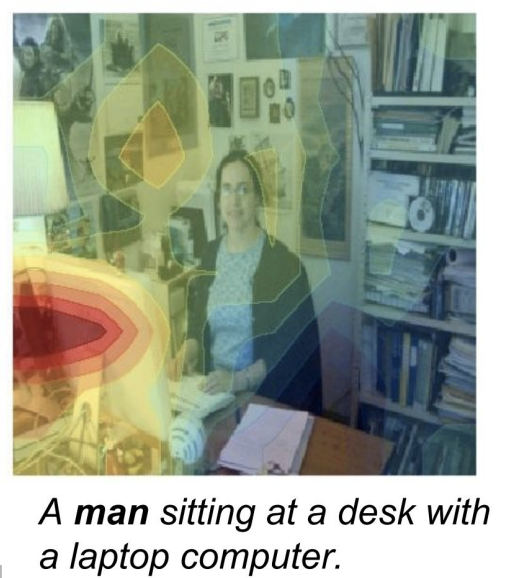}\end{center}
\section*{3. Our Approach}
To understand the gender bias present in Image captioning tasks, we decided to analyse the model of the paper Show, Attend and Tell [\cite{sat}], it being a popular recent work on Image captioning. We first conducted experiments to determine how much and in what form gender bias is present in the dataset and the model. This led us to realize that often the incorrect captions are a result of the relations learnt between certain activity/object words and gender words. Since the dataset has a biased distribution of gender-activity pairs, the model learns this bias. \\

As an approach to mitigate the effect of biased data distribution, we thought about techniques to make the data unbiased in order to avoid the model from observing any unwarranted relations. We thought that this could be done by evening out the samples for males and females for each activity. However, owing to the innumerable activities, and sparse occurrence of most scenes and activities, with certain rare activities having instances only with one gender; we realized that this would not be a feasible idea. Moreover, even if possible, this strategy of balancing the data would be very restrictive and require manual analysis and correction for every new dataset. So we decided to look for another direction.\\

Our goal was essentially to prevent the model from making strong priors about the activity based on the gender observed or vice versa. This was equivalent to asking the model to be gender neutral and only observe the image while predicting the activity in the image, and to only look at the person to predict the subject's gender, and not just relating the two based on priors. Though, relational priors should be helpful in strengthening a guess based on the image, and considering that the test distribution is the same as training distribution, should give better results. However, preventing unwarranted conclusions from these priors (that neglect the image) would be very challenging in this way. Also, we believe that for a given test image in real-time, the model should not correlate such ideas, since women and men should be considered equally capable of doing most activities, and thus be identified irrespective of the activity. Hence, we decided to decouple these two decisions.\\

So we decide to split our goal into two tasks, drawing from \cite{misra2016seeing}. One, a gender neutral image captioning task, and the other being just gender identification task in the image. For the first goal, we make the Show Attend and Tell model gender neutral by removing all instances of gender from the training data. For this we replace gender words like man, woman, men, lady, guy, etc. with person or people such that no gender specific relations are learned. Thereafter, we train the Gender Agnostic Show Attend and Tell network. For the second goal,  we used an existing popular model, trained to identify the gender of the humans present in the image. The results from both these tasks are then combined to obtain the final captions. Our proposed overall model is named - Show Attend and Identify - SAI, since it identifies the gender separately after obtaining a caption for the image.
\begin{center}\includegraphics[height=6cm]{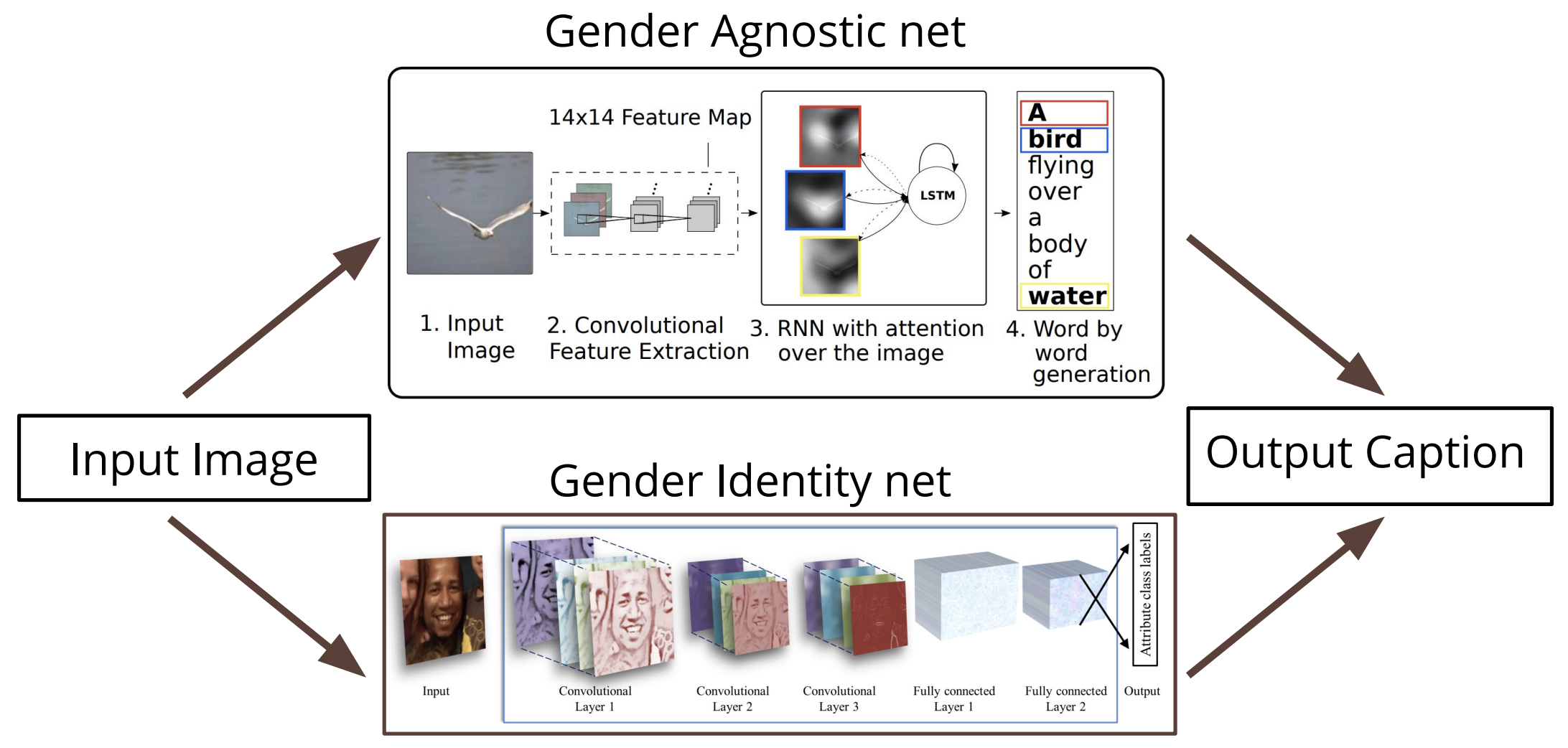}\end{center}
The above figure shows the top branch as the gender agnostic Show, Attend and Tell model and the bottom branch specialised in gender identification. We replaced the gender neutral words (person, people) with gender words (man, woman, men, women) based on the output of the Gender Identity net.
\section*{4. Datasets}
Our work is primarily on the MSCOCO dataset. To correctly evaluate the performance of our model, we split the data into subsets.  The following 3 subsets are used for experiments and evaluation of our model and comparison with the existing Show, Attend and tell model\cite{sat}.
\begin{itemize}
    \item \textbf{MSCOCO Gender confident dataset}: Because of the rampant human annotator bias infecting the data, there was a severe need for a dataset which is accurate and for which the labels are consistent. Since the test data is composed of total 40K instances and we wanted to test on a reasonable sized data, manual selection was not possible. Hence, drawing from \cite{women}, we relied on the fact that for a confident presence of a male or female in the image, all captions for that image should have a consistent mention of the gender. Hence, we picked up those image-caption instances for which all the captions had the same gender word (which can be either of man, woman, men, women, lady, ladies, guy, guys, etc.) This gave us a subset of 2036 images.
    \item \textbf{MSCOCO Human dataset}: We wanted to analyse the performance of our model to test the quality of captions generated for human-activity and human-object pairs and how that gets affected. Hence, we filtered those images for which any of the captions contain a human identifier word (which can be either of person, people, player, skier, snowboarder, man, woman, men, women, lady, ladies, guy, guys, etc.). This is  done by assuming that if a human being is present, atleast one of the captions would mention it. Also, if any of the captions mentions a human identity word, a human being must be present in that image. This gave us a dataset of around 19,051 images.
    \item \textbf{MSCOCO Nature dataset}: We wanted to analyse the performance of our model on images without the presence of any human beings to determine the performance and the effects, if any. Hence, we filtered out images where no human being is present and we relied on the fact that if none of the captions for an image contain a human label (which can be either of person, people, player, skier, snowboarder, man, woman, men, women, lady, ladies, guy, guys, etc.) then the image does not contain any human being. This gave us a subset of around instances 21,453 images
\end{itemize}

\section*{5. Experiments and Results}
For the gender Identificiation task, we use \cite{levi2015age} in our experiments.
We evaluated our model on the three datasets created and described above separately in order to observe the performance on each of the datasets. Since, our gender agnostic network provides predictions with person as the output, we could compare the overall quality of the captions predicted by the model by comparing the gender-neutral predictions. Hence, we also report results for this. For Show Attend and Tell, the predictions were made gender neutral (by replacing the words man, woman, girl, boy, men, etc with person or people). Ground-truth captions were also manually made gender neutral by replacing gender aware words with gender neutral ones.\\ \\ SAT is the Show Attend and Tell model. SAT-N denotes the comparison when the predicted captions of SAT are manually made gender neutral and compared with gender-neutral ground truth captions. SAI- Show, Attend and Identify is our complete model. SAI-N is the gender neutral component of our model. The following tables represent the BLEU, METEOR, ROUGE$_L$ and CIDEr scores obtained on the different datasets.

\begin{center}
\captionof{table}{Performance on confident dataset, both in gender aware and gender neutral settings (MSCOCO Gender Confident)} \label{tab:title1} 
 \begin{tabular}{||c c c c c c c c||} 
 \hline
 & Bleu1 & Bleu2 & Bleu3 & Bleu4 & METEOR & $ROUGE_L$ & CIDEr \\ [0.5ex]
 \hline\hline
SAT & 0.59 & 0.38 & 0.25 & 0.17 & 0.20 &0.49 &\textbf{0.44} \\ 
 \hline
 SAI & 0.59 & 0.38 & 0.24 & 0.16 & 0.19& 0.49 & 0.43\\
 \hline
SAT-N & 0.62 & 0.42 & 0.28 & \textbf{0.20} & \textbf{0.22} & 0.52 & \textbf{0.44}\\
\hline
SAI-N & \textbf{0.64} & \textbf{0.43} & \textbf{0.29} & 0.19 & \textbf{0.22} & \textbf{0.53} & \textbf{0.44}\\
\hline
\end{tabular}
\end{center}

\begin{center}
\captionof{table}{Performance on images with no humans (MSCOCO Nature)} \label{tab:title} 

 \begin{tabular}{||c c c c c c c c||} 
 \hline
  & Bleu1 & Bleu2 & Bleu3 & Bleu4 & METEOR & $ROUGE_L$ & CIDEr \\ [0.5ex] 
 \hline\hline
SAT & 0.62 & \textbf{0.41} & \textbf{0.27} & \textbf{0.18} & 0.20 &0.49 &\textbf{0.55} \\ 
 \hline
 SAI & 0.62 & 0.40 & 0.26 & 0.17 & 0.20& 0.49 & 0.54\\
 \hline
\end{tabular}
\end{center}

\begin{center}
\captionof{table}{Performance on images with humans (MSCOCO Human)} \label{tab:title} 

 \begin{tabular}{||c c c c c c c c||} 
 \hline
  & Bleu1 & Bleu2 & Bleu3 & Bleu4 & METEOR & $ROUGE_L$ & CIDEr \\ [0.5ex] 
 \hline\hline
SAT-N & 0.60 & 0.38 & \textbf{0.25} & \textbf{0.17} & 0.20 &\textbf{0.48} &0.47 \\ 
 \hline
 SAI-N & \textbf{0.61} & 0.38 & 0.24 & 0.16 & 0.20& 0.47 & 0.47\\
 \hline
\end{tabular}
\end{center}
A major limiting factor for the performance of our model is the accuracy of the gender identity CNN. It is a popular model with one of the best results reported. However surprisingly the model had a significantly low accuracy of around 50\% on the MSCOCO Confident dataset. This poses as a major roadblock.
\begin{center}\includegraphics[height=6cm]{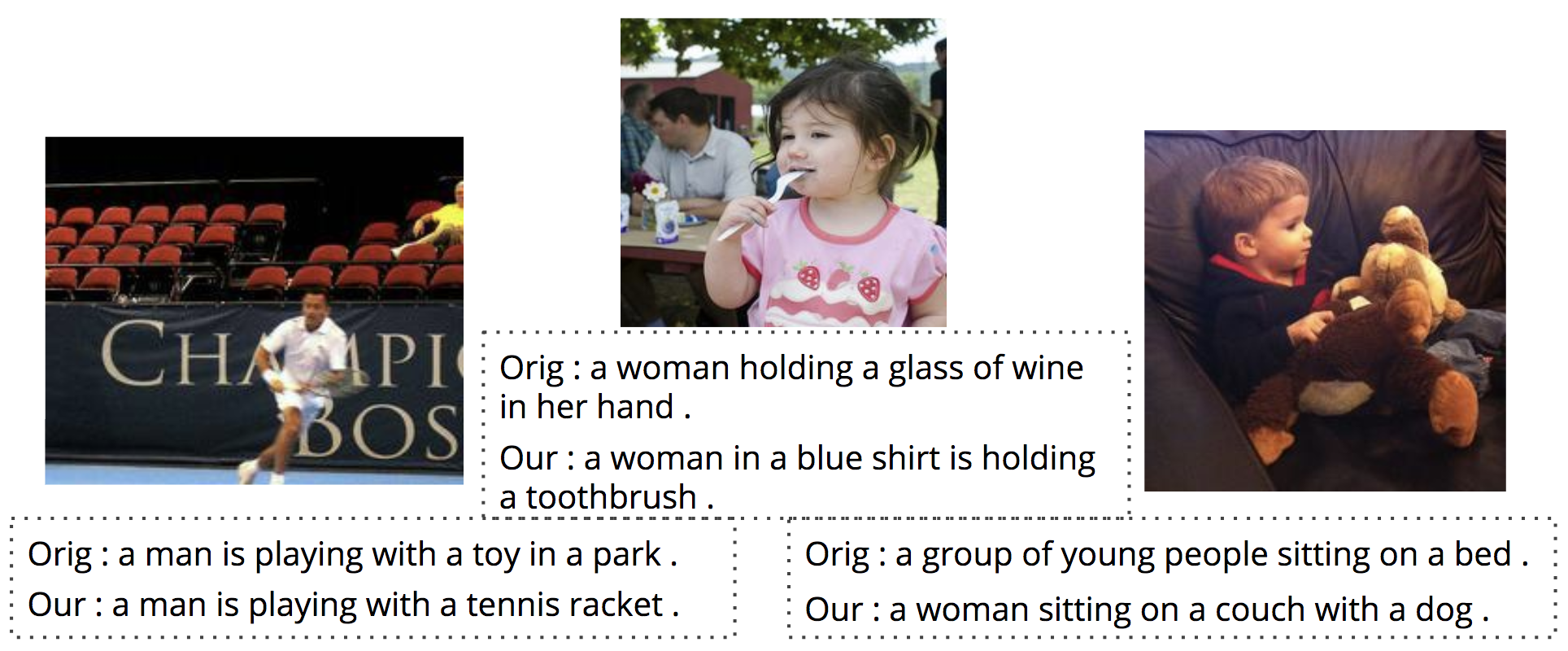}\end{center}
The above example shows qualitative results of our model and the show attend and tell model.
\section*{6. Conclusion and Discussion}
This project has demonstrated that gender bias is clearly prevalent in existing datasets and models. We propose a technique to do away with the bias in image captioning. Though our work is limited to images where all humans need to belong to the same gender, however it demonstrates that developing a gender neutral model allows us to get rid of the gender biased relations. The results depict that this leads to an improvement in the overall caption quality, while the performance for non-human images is not affected.  Also, we noticed that on removing man/woman from training data, our gender neutral model predictions saw a sudden emergence of the words 'male' and 'female'. This depicts the close relation between the word embeddings for these. Overall, we believe that the approach to get unbiased models involves having the decoupling procedure to avoid any gender specifications and thus unwarranted relations from entering the model.
\section*{7. Problems and Future Work}
We started this project with the task of removing gender bias from VQA models, however VQA at present remains one of the most challenging problems, having been called the 'visual turing test' by some. Hence, most VQA models at present are too naive making the task all the more challenging. Hence, we worked on the removal of gender bias from Image captioning tasks in which decent results have been obtained by some of the state-of-the-art methods including Show, Attend and Tell which we use as our basenet. One could experiment with the use of debiased word  embeddings to further remove gender bias from the model. One could learn a gender identification network which not only identifies male/female but also has a third category of person, which should be chosen in cases when the discrimination is not possible. More work on the gender network could lead to improvements for our overall model. 



\section*{\instructions{Bibliography}}
\instructions{Do not  forget to include bibliographic references. You need to create your own .bib file. If you call it {\tt mybib.bib}
  and put it in the same directory as this {\tt .tex} file, add
  {\tt$\backslash$bibliography\{mybib\}} before
  {\tt$\backslash$end\{document\}}.
This template uses the natbib package, which allows you use a citation
format that includes author names and years (instead of numerical
references).  You can use the {\tt$\backslash$cite\{\}} command (
``\cite{Mins:69} presented a damaging critique of the perceptron'')
or {\tt$\backslash$citep\{\}} (``A damaging critique of the perceptron
\citep{Mins:69} had long-lasting effects on the development of AI'').  
(NB: for a lot of papers on the ACL anthology, you can get the BibTeX
entry by changing the .pdf extension on the URL to  .bib,
e.g. from \url{http://aclweb.org/anthology/P/P17/P17-1001.pdf} to
\url{http://aclweb.org/anthology/P/P17/P17-1001.bib})
}
\bibliographystyle{plainnat} 
\bibliography{mybib}
\end{document}